\documentclass[10pt,twocolumn,letterpaper]{article}

\usepackage{iccv}
\usepackage{times}
\usepackage{epsfig}
\usepackage{graphicx}
\usepackage{amsmath,amssymb,amsfonts,amsthm}

\usepackage[pagebackref=true,breaklinks=true,letterpaper=true,colorlinks,bookmarks=false]{hyperref}

\iccvfinalcopy 

\def\iccvPaperID{7836} 

\ificcvfinal\fi

\begin{document}

\title{Salient Object Ranking with Position-Preserved Attention}

%

\author{Hao Fang\\
Zhejiang University\\
Alibaba Group\\
{\tt\small fanghao\_zju@163.com}
\and
Daoxin Zhang\\
Alibaba Group\\
{\tt\small daoxin.zdx@alibaba-inc.com}
\and
Yi Zhang\\
Zhejiang University\\
{\tt\small 21921091@zju.edu.cn}
\and
Minghao Chen\\
Zhejiang University\\
{\tt\small minghaochen01@gmail.com}
\and
Jiawei Li\\
Alibaba Group\\
{\tt\small mingong.ljw@alibaba-inc.com}
\and
Yao Hu\\
Alibaba Group\\
{\tt\small yaoohu@alibaba-inc.com}
\and
Deng Cai\\
Zhejiang University\\
{\tt\small dengcai78@qq.com}
\and
Xiaofei He\\
Zhejiang University\\
{\tt\small xiaofei\_h@qq.com}

}

\maketitle
\ificcvfinal\thispagestyle{empty}\fi

\begin{abstract}
Instance segmentation can detect where the objects are in an image, but hard to understand the relationship between them. We pay attention to a typical relationship, \textbf{relative saliency}. A closely related task, salient object detection, predicts a binary map highlighting a visually salient region while hard to distinguish multiple objects. Directly combining two tasks by post-processing also leads to poor performance. There is a lack of research on relative saliency at present, limiting the practical applications such as content-aware image cropping, video summary, and image labeling.

In this paper, we study the Salient Object Ranking (SOR) task, which manages to assign a ranking order of each detected object according to its visual saliency. We propose the first end-to-end framework of the SOR task and solve it in a multi-task learning fashion. The framework handles instance segmentation and salient object ranking simultaneously. In this framework, the SOR branch is independent and flexible to cooperate with different detection methods, so that easy to use as a plugin. We also introduce a Position-Preserved Attention (PPA) module tailored for the SOR branch. It consists of the position embedding stage and feature interaction stage. Considering the importance of position in saliency comparison, we preserve absolute coordinates of objects in ROI pooling operation and then fuse positional information with semantic features in the first stage. In the feature interaction stage, we apply the attention mechanism to obtain proposals' contextualized representations to predict their relative ranking orders. Extensive experiments have been conducted on the ASR dataset. Without bells and whistles, our proposed method outperforms the former state-of-the-art method significantly. The code will be released publicly available on https://github.com/EricFH/salient\_object\_ranking\_with\_PPA.

\end{abstract}

\begin{figure}[!tb]
\centering
\includegraphics[width=0.48  \textwidth]{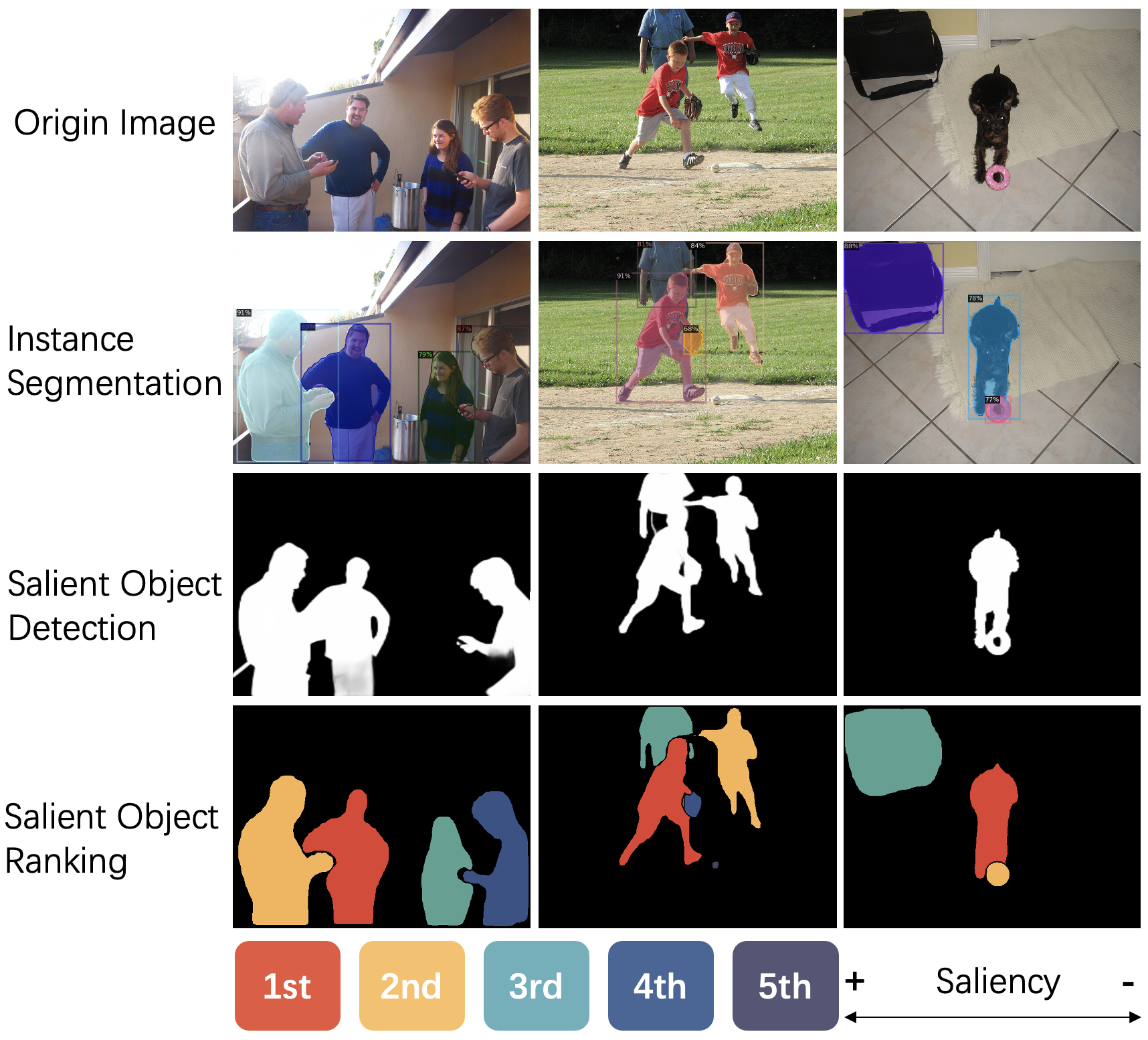}
\setlength{\belowcaptionskip}{-1.5cm}   
\caption{The salient object ranking (SOR) task assigns a ranking order to each detected object according to their visual saliency. Instance segmentation can detect objects but can not obtain the relationship between them. In the meantime, salient object detection can highlighting the most attractive regions but can not distinguish them. (Best viewed in color.)}
\label{fig:head}
\end{figure}

\section{Introduction}
Instance segmentation has made tremendous progress in recent years~\cite{he2017mask,lee2020centermask}. To get a deeper understanding of images, exploring the relationship between objects after detecting their locations is meaningful for researchers. A typical relationship is relative saliency that compares which one is more attractive than another. Salient Object Detection (SOD) is a closely related task aiming to locate regions where attract human visual attention. Most works formulate this task as a pixel-wise binary prediction task~\cite{zhang2017amulet, zhang2017learning,he2017delving,hou2017deeply,wang2016saliency,wang2017stagewise,hu2017deep,li2016deep,lee2016deep,liu2016dhsnet,najibi2018towards,zhang2016unconstrained}. Since SOD predicts all salient areas at pixel-level rather than instance-level, it has limitations distinguishing multiple objects in real scenes. (shown in Fig. \ref{fig:head})

\begin{figure*}[!t]
\centering
\includegraphics[width=0.9 \textwidth]{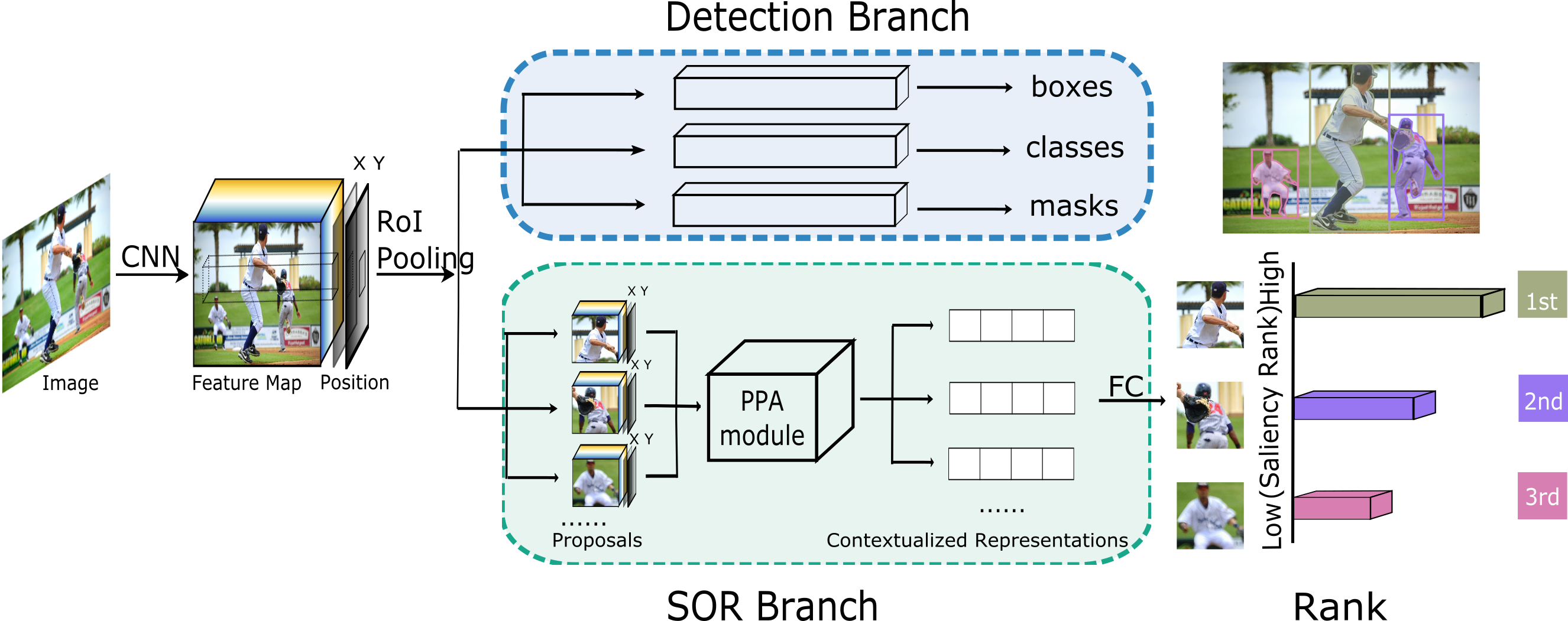}
\caption{The proposed end-to-end salient object ranking (SOR) architecture is based on a multi-task learning framework. Semantic features from the shared feature extractor are concatenated with positional coordinate maps before ROI pooling. Then detection branch (shown in the blue box) predicts instance segmentation results. The SOR branch (shown in the green box) combines semantic features with position embedding and interacts features between proposals. It employs a novel Position-Preserved Attention (PPA) module to get contextualized representations and makes the final ranking prediction via FC layers.}
\label{arch}
\end{figure*}
Salient Object Ranking (SOR) is a recently proposed problem introduced by~\cite{islam2018revisiting} handling scenes of multiple objects. It assigns a unique ranking order to every detected object according to its visual saliency. The salient ranking orders of objects in an image reflect human attention shifting process~\cite{siris2020inferring}, which helps researchers explore how humans interpret an image. In the meantime, substantial downstream applications are in huge demand of SOR. Representative ones, such as content-aware image cropping~\cite{zhu2021horizontal,chen2016automatic}, image parsing~\cite{7780764,tighe2013finding}, and image captioning~\cite{xu2015show,yao2018exploring}, can not be well solved by employing current object detection and SOD methods.

Works on the SOR task are limited. We can categorize them into FCN-based~\cite{islam2018revisiting,wang2019ranking} and Detection-based~\cite{siris2020inferring}. FCN-based methods predict saliency ranking orders pixel by pixel as SOD does. Pixels in the same instance could be predicted to different ranking orders. It does not meet the requirement of SOR, which aims at assigning the same ranking order for the same object. Although complicated post-processing cooperated with other detection models can relieve this issue, the performance is unstable. Siris \etal~\cite{siris2020inferring} proposes a Detection-based method. It first trains a detector extracting features of each proposal. Then it combines top-down and bottom-up information with proposals' features to predict their ranking orders. However, this network can not be trained end-to-end. Detection loss and SOR loss are hard to be optimized jointly. The methods mentioned above do not fully utilize positional information, which is a significant factor in ranking objects' saliency. That is to say, an object in the center tends to be more attractive than one in the corner. Also, an object with a larger scale is usually more eye-catching than a smaller one. Another essential factor is the correlation between objects. A more attractive object will lower the visual saliency of others. (shown in Fig. \ref{sentitive})


In this paper, we propose an end-to-end framework for the first time of the SOR task and solve it in a multi-task learning fashion. In this framework, the detection and salient object ranking branches are parallel rather than sequential. We can optimize SOR loss and detection loss jointly to achieve better performance. The SOR branch completes an independent ranking prediction task. So it could be considered as a flexible plugin with diverse detection methods.

We further introduce the Position-Preserved Attention (PPA) module tailored for the SOR branch. PPA consists of the position embedding stage and the feature interaction stage. In the position embedding stage, besides semantic features extracted from ROI pooling~\cite{ren2015faster,he2017mask}, positional information of each object is considered. Both absolute positions in the image and the relative position between each other help rank objects' saliency. However, the common ROI pooling operation will crop object-level features from the whole feature map and lost objects' positional information. To address this issue, we concatenate positional coordinate maps with the whole feature map before ROI pooling. Then we pass them into ROI pooling together. This position-preserved pooling process finally obtains the corresponding positional information of each object. After fusing semantic features and position embedding, we get richer features of each object. 

Since SOR aims at obtaining relative saliency ranking between each other rather than a specific salient label, the feature interaction stage is of vital importance. In this stage, the attention mechanism is utilized to make objects receive other objects' features and obtain contextualized representations to predict their relative ranking orders. We employ the encoder of Transformer~\cite{vaswani2017attention} to implement the attention mechanism. Each object-level feature is considered as a \textit{visual token}, which is the input of the encoder of Transformer.

Extensive experiments have been conducted on the ASR dataset~\cite{siris2020inferring}, a recently proposed salient object ranking dataset. Without bells and whistles, our method outperforms the former state-of-the-art method significantly.

In summary, the main contributions of this work include:
\begin{itemize}
\item We propose the first end-to-end framework of the SOR task and solve it in a multi-task learning fashion. We can optimize SOR loss and detection loss jointly to achieve better performance. The SOR branch is flexible to cooperate with other detection methods.
\item We introduce a Position-Preserved Attention (PPA) module tailored for the SOR branch, which preserves absolute coordinates of objects in ROI pooling operation and then fuses positional information with semantic features. In the feature interaction stage, an attention mechanism is applied between each object to obtain contextualized representations.
\item Our method outperforms the former state-of-the-art method significantly on the ASR dataset. It can serve as a strong baseline to facilitate future research on SOR.
\end{itemize}


\section{Related Work}

\subsection{Salient Object Detection}

As an essential problem in the computer vision community, salient object detection (SOD) has attracted many researchers' attention in recent years. The majority of SOD methods~\cite{borji2019saliency,wang2021salient} are designed to detect visually salient regions and the task is formulated as a pixel-wise binary prediction problem. Early works try to construct heuristic features in pixel-level~\cite{itti1998model,ma2003contrast}, patch-level~\cite{liu2010learning,achanta2008salient}, and region-level~\cite{cheng2014global}. With the advance of Convolutional Neural Networks (CNNs), features learned by CNN are leveraged to improve the performance of SOD. SOD methods can be roughly categorized into three fashions, \textit{i.e.}, super-pixel based~\cite{hu2017deep,lee2016deep}, coarse-to-fine~\cite{zhang2017learning,liu2016dhsnet}, and method utilizing bottom-up and top-down pathways~\cite{luo2017non,hou2017deeply,zhang2017amulet}.

\subsection{Salient Object Ranking}
Salient Object Ranking is a new proposed problem introduced by~\cite{islam2018revisiting} in 2018. 
Islam \etal~\cite{islam2018revisiting} first formulates the SOR problem and proposes an FCN-based model applying hierarchical representation of relative saliency and stage-wise refinement. Li \etal~\cite{li2014secrets} finds that there is a strong correlation between fixations prediction (FP) and salient object detection (SOD). Wang \etal~\cite{wang2019ranking} also proposes a SOR model on a video dataset leveraging FP and SOD branches. These methods finally predict a saliency ranking map, which could be seen as the post-process of SOD. A detection-based method is proposed by~\cite{siris2020inferring}, which first pre-trains a detector to extract object-level features. Then SAM and SMM modules are applied to fuse global features and mask features with object features. Finally, a simple classifier is followed to predict ranking orders. This method is object-aware compared with former works, however, it needs two-stage training and is not end-to-end, which leads to a problem of optimizing detection loss and SOR loss jointly. Meanwhile, positional information is not considered and each ranking order is predicted independently without interaction.

\subsection{Visual Transformer}


Transformer is a neural network mainly based on the self-attention mechanism, which is first applied to natural language processing (NLP) tasks and has achieved significant improvements~\cite{vaswani2017attention,devlin2018bert,brown2020language}. There is an increasing number of works applying Transformer to computer vision tasks.
Chen \etal~\cite{chen2020generative} trains a generative model to auto-regressively predict pixels. ViT~\cite{dosovitskiy2020image} applies a pure transformer directly to sequences of image patches on classification. Transformer has also been utilized to address various computer vision problems, such as object detection~\cite{carion2020end}, semantic segmentation~\cite{wang2020max}, video processing~\cite{zeng2020learning}, and pose estimation~\cite{huang2020hand}. These works employ the Transformer on image-level or patch-level, while our method employs it on ROI-level (Region of Interest).


\section{Method}
This section presents the details of our proposed model, including the design motivation (Section \ref{sec:motiv}), the overall network architecture (Section \ref{sec:frame}), the Position-Preserved Attention module (Section \ref{sec:ppa}), and discussion on different position embedding schemes (Section \ref{sec:pos_emb}). We propose the first end-to-end framework of the SOR task and solve it in a multi-task learning fashion. In this framework, a CNN first extracts a shared feature map from an input image. Then coordinate maps of the X-axis and Y-axis are concatenated with feature maps before ROI pooling. After obtaining proposals with object-level features and positional information, we pass these proposals into the SOR branch and detection branch simultaneously to get final results.

\begin{figure}[!tb]
\centering
\includegraphics[width=0.48 \textwidth]{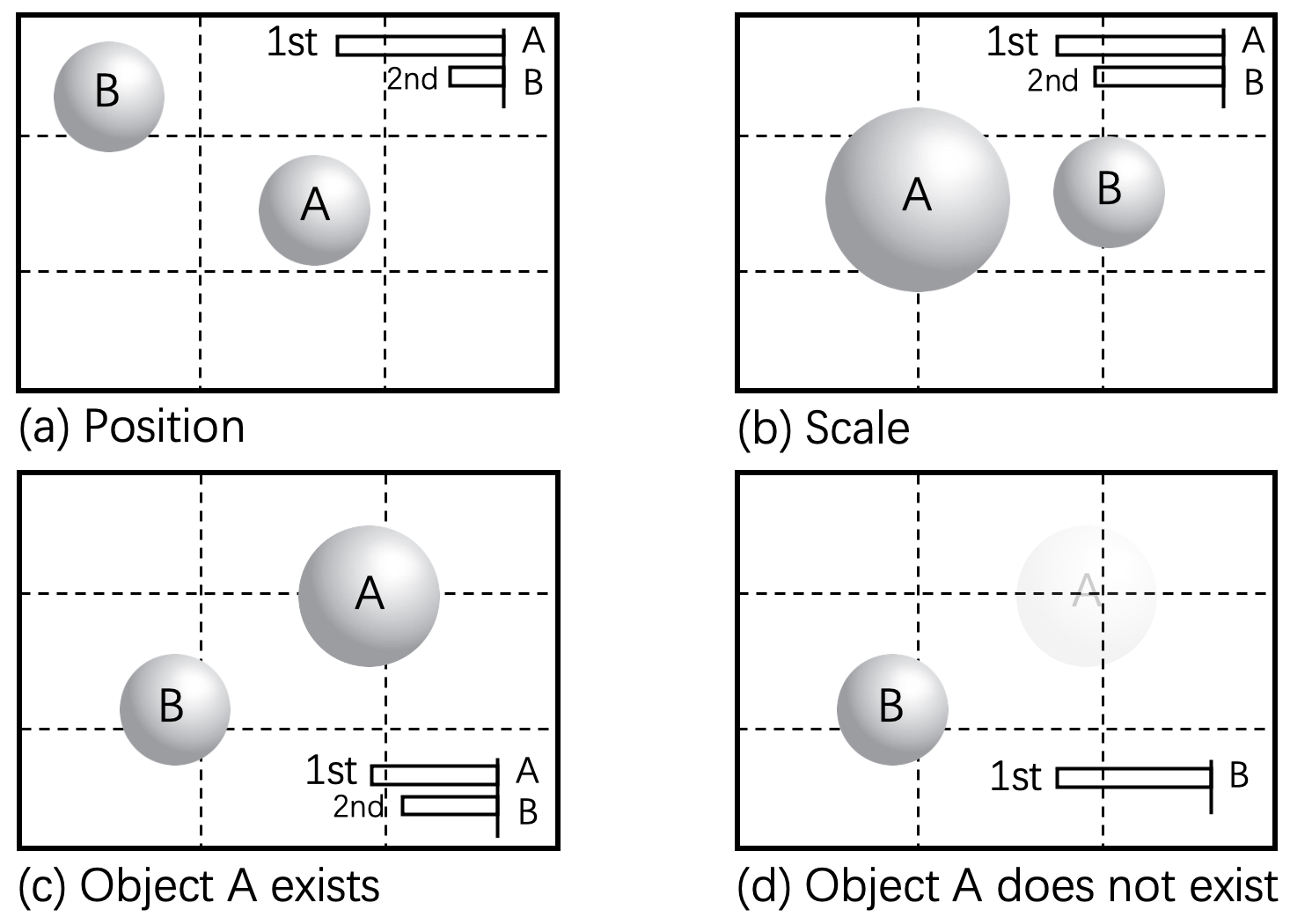}
\caption{Illustration of crucial factors that influence object salient ranking other than semantic information, where the histograms represent the SOR within corresponding images. In the figure, (a) and (b) show the position and scale influence saliency ranking respectively. (c) and (d) show the existence of object A will affect the ranking order of object B.}
\label{sentitive}
\end{figure}

\subsection{Motivation}
\label{sec:motiv}
Compared with common object detection and segmentation tasks, one of the distinctive characteristics of the SOR task is its sensitivity to position and scale. In other words, the position and scale of an object in the image do not affect the character's nature (category) and appearance characteristics (detection and segmentation). This information is usually regarded as the prior knowledge of the task, which is absolute information. Nevertheless, the SOR task is different. As shown in Fig. \ref{sentitive} (a), the same object in the center or corner of the image may directly change from the foreground subject to the background. Similarly, the scale also has an important influence on this saliency ranking task, which is demonstrated in Fig. \ref{sentitive} (b).

The other difference is that the ranking order of one object is affected by other objects' existence. The existence of one more attractive object will lower the saliency of other objects. For instance, when comparing Fig. \ref{sentitive} (c) with Fig. \ref{sentitive} (d), we find that the removal of object A influences the salient ranking of object B. Therefore, feature interaction between objects is also an important step of this task, that is, features of objects are relative in the SOR task. How to effectively utilize positional information and mutual information is the key to solve this task, which is difficult to tackle by applying the detection framework directly.

\subsection{Multi-Task Learning Architecture}
\label{sec:frame}
The overall network architecture is shown in Fig. \ref{arch}. The framework is composed of three components:
\begin{enumerate}
    \item Backbone: A commonly used CNN network is applied as a feature extractor. The input is a raw image, while the output is a feature map. ROI pooling operation will be applied to crop object-level features of each proposal. To add positional information to our Position-Preserved Attention Module, we concatenate coordinates of the X-axis and Y-axis with feature map before ROI pooling: $[FeaMap;PosMap]$\footnote{where $[\cdot;\cdot]$ denotes concatenating along the channel dimension}. We will discuss more details in Section \ref{sec:ppa}.
    
    \item Detection branch: Off-the-shelf detection methods can be used in this branch, such as~\cite{he2017mask,lee2020centermask}. The target of this branch is to detect objects and predict their locations, classes, and masks. The positional information of each proposal is not used in this branch.
    
    \item SOR branch: SOR branch is designed to assign each proposal a ranking order according to their visual saliency. The goal of the SOR branch is ranking proposals rather than detecting their existence. PPA module plays a major role in this branch, which is comprised of position embedding stage and feature interaction stage. In the first stage, semantic information and positional information are fused to obtain \textit{visual tokens}. Then they are passed into the feature interaction stage and get contextualized representations for each proposal. Finally, a fully connected layer is followed to predict each proposal a ranking order.

\end{enumerate}
\paragraph{Loss Function} We define our training loss function as follows: 
\begin{align}
\mathcal{L} = \mathcal{L}_{det} + \lambda \mathcal{L}_{sor},
\end{align}
where $\mathcal{L}_{det}$ is the detection loss, for instance, $\mathcal{L}_{det}=\mathcal{L}_{box}+\mathcal{L}_{cls}+\mathcal{L}_{mask}$. The bounding box loss $\mathcal{L}_{box}$, classification loss $\mathcal{L}_{cls}$, and mask loss $\mathcal{L}_{mask}$ are identical as those defined in MaskRCNN~\cite{he2017mask}. There could be some differences in the details of $\mathcal{L}_{det}$ when the framework applies diverse detection methods~\cite{he2017mask,lee2020centermask}. $\mathcal{L}_{sor}$ is the SOR loss, which is the cross-entropy loss between the distribution of predicted ranking order and ground-truth ranking order. In all experiments, we set $\lambda$ to $1.0$.

\subsection{Position-Preserved Attention Module}
\label{sec:ppa}

To solve the aforementioned issues in Section \ref{sec:motiv}, we propose the Position-Preserved Attention (PPA) module. It is the major part of the SOR branch, which consists of the position embedding stage and feature interaction stage. The position embedding stage enriches the semantic features with positional information, while the feature interaction stage utilizes mutual information between proposals.

\begin{figure}[!tb]
\centering
\includegraphics[width=0.5 \textwidth]{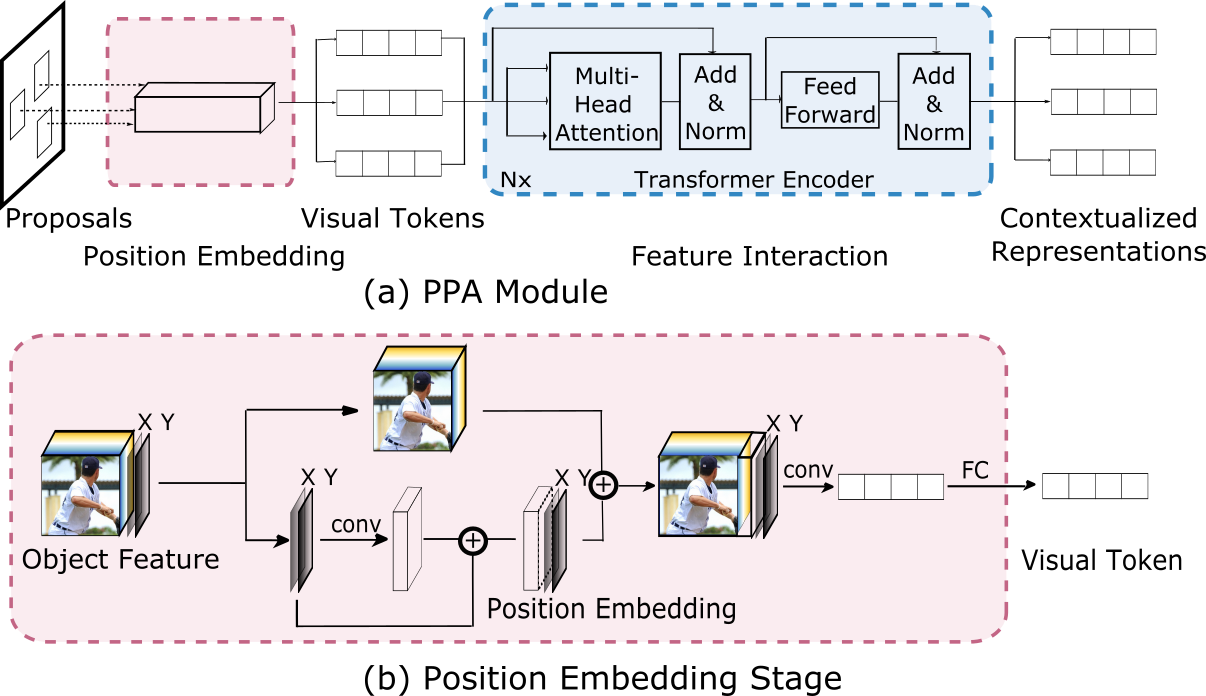}
\caption{
PPA module is the center of the SOR branch. It consists of position embedding stage (red block) and features interaction stage (blue block) specifically. The position embedding stage enriches the semantic features with positional information and outputs vectorized \textit{visual tokens}. The feature interaction stage utilizes mutual information between proposals by means of a Transformer encoder~\cite{vaswani2017attention}. The PPA module finally outputs contextualized representations of each proposal for ranking prediction.}
\label{ppa}
\end{figure}

The input of the PPA module is the proposals' features with position (\textit{e.g.} $N\times 14 \times 14 \times (256+2)$, $N$ denotes the number of proposals, $14$ is the ROI pooling size,
the number of channels of feature map and position indexes are $256$ and $2$, respectively). For the i-th proposal with BBox coordinates $bbox_i$, its feature after RoI pooling is $[fea_i;pos_i] = \mathrm{RoIPooling}([FeaMap;PosMap], bbox_i)$. The outputs of the PPA module are contextualized representations of each proposal (\textit{e.g.} $N\times 1024$). The detailed structure is shown in Fig. \ref{ppa} (a).

\paragraph{Position embedding stage}
This stage is shown in Fig. \ref{ppa} (b), which aims at fusing semantic features and positional information for each proposal. Firstly, the feature map is divided into the semantic part and positional part. Then a convolution layer with ReLU activation function is applied on the position part to extract low-level features: $pos\_fea_i = \mathrm{Conv}(pos_i)$. 
Original position and low-level feature are concatenated together, and position embedding of the proposal is obtained: $pos\_embeding_i = [pos_i; pos\_fea_i]$. Then both semantic feature and position embedding are concatenated and passed following four convolution layers together:  $fea_i = \mathrm{Convs}([fea_i; pos\_embeding_i])$. The fused feature map is flattened and finally transformed into a vector with $1024$ channels after two fully connected layers. Each proposal is transformed as a 1-D vector respectively, which is considered as a \textit{visual token}. The concept of the \textit{visual token} is borrowed from NLP, as we use the encoder of Transformer in the following stage.

\paragraph{Feature interaction stage}
To utilize mutual information between proposals, we apply the encoder of Transformer~\cite{vaswani2017attention}, which is benefited from the self-attention mechanism. We follow the standard Transformer encoder structure, which is demonstrated on the right side of Fig. \ref{ppa} (a). It consists of alternating layers of multi-head self-attention and feed-forward neural network (FFNN) blocks. Layer normalization (LN)~\cite{ba2016layer} is applied after each block and residual connections~\cite{wang2019learning}. GELU~\cite{hendrycks2016gaussian} is used as an activation function in FFNN.


\subsection{Position Embedding Discussion}
\label{sec:pos_emb}
\begin{figure}[!tbh]
\centering
\includegraphics[width=0.48 \textwidth]{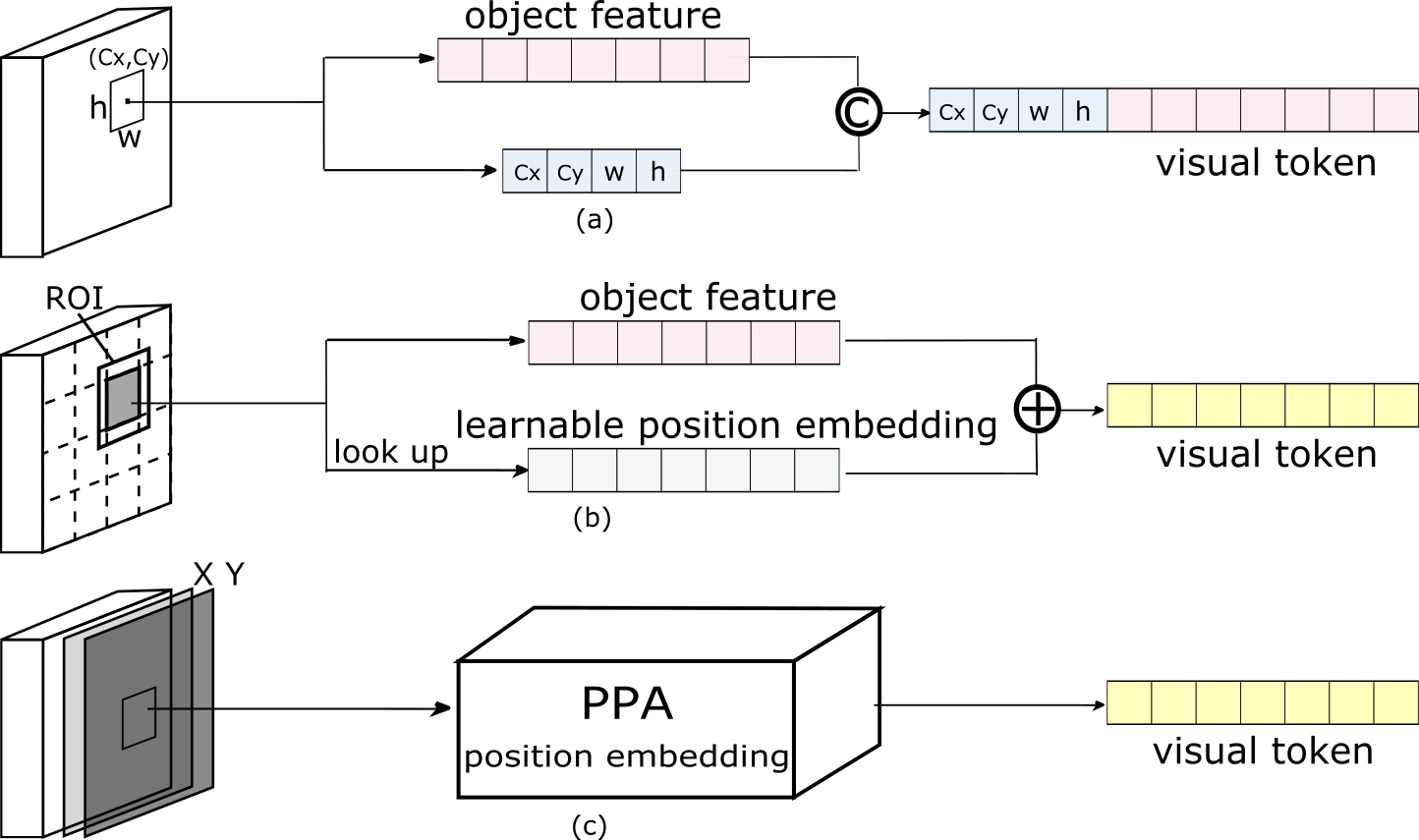}
\caption{Three position embedding schemes. (a) directly concatenate normalized [$C_x$,$C_y$,$w$,$h$] to ROIs' features. (b) add learnable look-up position embedding to features. (c) PPA position embedding stage.}
\label{pos}
\end{figure}

We have explored different position embedding schemes since sensitivity to position and scale is a key factor for the SOR task. We start from the simple concatenation method shown in Fig. \ref{pos} (a). Proposals' features are cropped from the feature map after the ROI pooling operation. Then we apply four convolution layers and then flattened them to 1-D vectors. After two fully connected layers, finally, object features are obtained. In this scheme, we simply concatenate object features, center coordinates $(c_x, c_y)$ and scales$(w, h)$ (normalized by the image width and image height) of responding bounding boxes to get \textit{visual tokens}.

Following position embedding methods used in vision transformers~\cite{dosovitskiy2020image,he2021transreid}, we try the 1-D learnable position embedding vector shown in Fig. \ref{pos} (b). This scheme will predefine fixed-number cells, each of which is corresponding a learnable embedding vector. However, the positions and scales of proposals, which are calculated rather than predefined, are real numbers and not enumerable. To address this issue, we use a quantitative method and grid the value space. 
Given a image with $W$ width and $H$ height, we first grid the space into $q \times q$ cells whose shape is $\frac{W}{q} \times \frac{H}{q}$. For a proposal's bounding box $bbox$ with $(C_x, C_y)$ center position, we calculate the index ($Id_x$, $Id_y$) of the proposal: $Id_x = \lfloor \frac{C_x \cdot q}{W} \rfloor, Id_y = \lfloor \frac{C_y \cdot q}{H} \rfloor$, which cell it belongs to. We use $Id_x$ and $Id_y$ to obtain index of predefined $Embeddings$ in $Id = Id_x \cdot q + Id_y\label{idx}$. Object features are got in the same way of the first scheme. Both object features and position embeddings are added together to obtain the final \textit{visual tokens}. Scale information can also be embedded in this way.


However, in this method, two proposals in close position will be grouped into the same cell. Quantitative error will be introduced and positional information will be inevitably lost. Using larger quantitative number $q$ can ease this error but causes more learnable parameters. Larger $q$ will also bring risk to the model. If the dataset is not large enough and some indexes of $Embeddings$ are not well trained, the corresponding position $bbox$ will have a bad representation in inference stage. The quantitative number $q$ will be a hyper-parameter difficult to tune. In experiments, the performance of $q=4$ and $q=8$ are comparable.

Compared with the above two schemes, our method, shown in Fig. \ref{pos} (c), attaches absolute positional information to the feature map before ROI pooling. This position-preserved process utilizes absolute positional information directly. Corresponding coordinate map of each proposal is preserved after ROI pooling. Then the proposal's feature and positional information are fused together in the following layers, rather than learned independently compared with the scheme (b). Our experiments show that the model gains more benefit in PPA's position embedding fashion.

\section{Experiments}

\subsection{Dataset}
We conduct experiments on the ASR dataset~\cite{siris2020inferring}, which is the only public released SOR dataset as far as we know. ASR dataset is a large-scale salient object ranking dataset by combining the MS-COCO dataset~\cite{lin2014microsoft} with the SALICON dataset~\cite{jiang2015salicon}. It consists of 78 object categories, and the average number of objects per image in the dataset is around 11. The annotation of instances is the same of MS-COCO dataset and the extra information is ranking order. In each image, the top-5 most visually salient instances own a unique ranking order ranging from 1 to 5, and other instances are considered as background.  The dataset is randomly split into 7646 training, 1436 validation, and 2418 test images, respectively.

\subsection{Evaluation Metrics}
We adopt evaluation methods as~\cite{siris2020inferring,islam2018revisiting} does to make a fair comparison, \textit{i.e.}, Salient Object Ranking (SOR) and Mean Absolute Error (MAE).
SOR metric calculates \textit{Spearman's Rank-Order correlation} between predicted ranking orders and the ground-truth ranking orders of salient objects. SOR indicates the correlation between two ranking order lists, and higher SOR means a higher positive correlation. To make it more interpretable, the SOR score is usually normalized to $[0,1]$. However, if there are no common salient objects between ground-truth and prediction, SOR is not suitable to measure performance in this case. To solve this problem, we do not take images into account where ground-truth objects have no overlap with predicted instances. The number of images we use to calculate is called \textit{Images used}. The more \textit{Images used}, the more reliable the SOR is. At the same time, more \textit{Images used} indicates better detection performance. 

MAE metric compares the average absolute per-pixel difference between the prediction saliency map and ground-truth map. Compared with the SOR metric, which focuses on ranking order, MAE takes both detection results and ranking results into consideration.

\subsection{Implementation Details}





We take an end-to-end training strategy. The shape of input images is $640 \times 480$, which is the same as the original images. To explore the model's capability itself, we do not introduce any data augmentation tricks. We apply SGD~\cite{bottou2010large} as an optimizer with momentum 0.9 and gamma 0.1, and the base learning rate is set to 1e-4. We utilize the warm-up~\cite{goyal2017accurate} strategy in the first 1000 iterations and apply a multi-step policy whose weight decay factor is 0.1. We mainly use VoV-39~\cite{lee2019energy} as backbone and CenterMask~\cite{lee2020centermask} as detector in experiments unless otherwise specified. We set the mini-batch size to 16 and train the network for 54000 iterations. All models are implemented with PyTorch on 2 TITAN RTX GPUs.


The inference stage is a sequential process. We first take the object with the highest score in rank 1 class as the top 1 salient object. Then we remove this object from candidate objects and choose the object with the highest score in the rank 2 class, and so on. In this way, we can obtain the top-5 salient objects, avoiding the case that multiple objects are assigned to the same ranking order. 

\subsection{Main Results}

\begin{table}[!tbh]

\begin{center}

\begin{tabular}{lccc}

\hline
Method & MAE$\downarrow$ & SOR$\uparrow$ & \#Imgs used$\uparrow$ \\
\hline
RSDNet~\cite{islam2018revisiting} & 0.139 & 0.728 & \textbf{2418}          \\
S4Net~\cite{fan2019s4net}  & 0.150 & \textbf{0.891} & 1507          \\
BASNet~\cite{qin2019basnet} & 0.115 & 0.707 & 2402         \\
CPD-R~\cite{wu2019cascaded}  & \underline{0.100} & 0.766 & 2417        \\
SCRN~\cite{wu2019stacked}   & 0.116 & 0.756 & \textbf{2418}          \\
\hline
ASRNet~\cite{siris2020inferring} & 0.101 & 0.792 & 2365          \\
Ours   & \textbf{0.081} & \underline{0.841} & 2371       \\
\hline
\end{tabular}

\end{center}
\caption{Comparison with state-of-the-art methods on the ASR dataset. 
The first five methods only provide a single binary saliency map without object segmentation. ASRNet and our methods predicts instance segmentation maps. $\uparrow$($\downarrow$) means the higher (lower) the better. The bold number is the top score and the underlined number is the second.}
\label{main table}
\end{table}

\begin{figure*}[!t]
\centering
\includegraphics[width=1 \textwidth]{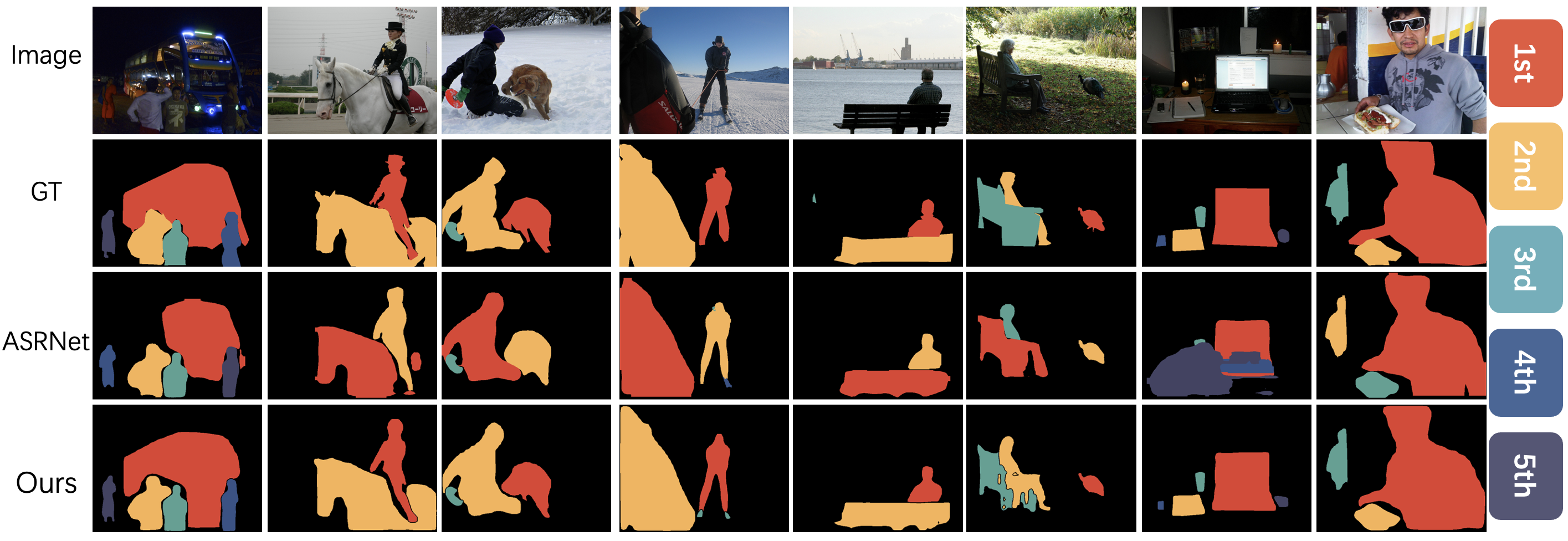}
\caption{Results on the ASR dataset with ground-truth, ASRNet and our model. Our proposed method improves both instance segmentation and ranking quality compared to state-of-the-art method ASRNet~\cite{siris2020inferring}.}
\label{vis}
\end{figure*}

\paragraph{Quantitative Evaluation} We make a comparison between our method and six state-of-the-art methods, including RSDNet~\cite{islam2018revisiting}, S4Net~\cite{fan2019s4net}, BASNet~\cite{qin2019basnet}, CPD-R~\cite{wu2019cascaded}, SCRN~\cite{wu2019stacked} and ASRNet~\cite{siris2020inferring}. The RSDNet first introduces the saliency object ranking problem. Note that the first five methods only provide a single binary saliency map instead of object segmentation. ASRNet predicts instance segmentation maps, but it needs two-stage training. We propose an end-to-end network for the first time.

Since the first five methods mentioned above have different outputs with ASRNet and our method. To make a fair comparison, we apply the same post-processing as ASRNet does, to get distinct saliency ranking orders for these models. For the S4Net, the original output is modified in order to predict up to 6 classes for each object instead of binary prediction. For the rest compared models (RSDNet, BASNet, CPD-R, SCRN), the predicted saliency ranking orders of ground-truth objects are calculated by averaging the pixel saliency values.

The experimental results are shown in Table \ref{main table}, which denotes our method outperforms the other methods mentioned before and achieves state-of-the-art results. Our model obtains best overall performance with better scores among all metrics (MAE, SOR, and \textit{Images used}). Even though RSDNet and SCRN models own higher \textit{Images used}, their single binary saliency maps contain many false saliency instances, which ensure these methods can cover the most salient objects. As a result, their SOR scores are comparatively lower than our method. S4Net obtains the highest SOR score, yet there are only two-thirds of test images used to calculate the SOR score. The rest images are ignored because their predicted saliency map can not match the ground-truth. It causes S4Net to suffer from the highest MAE and lowest \textit{Images used}. Since there is only one previous work that solves the SOR problem directly (\textit{i.e.} ASRNet), we should pay more attention to comparing ASRNet with our model. Results in the last two rows of Table \ref{main table} show that our model surpasses ASRNet significantly for all evaluation metrics, which suggests our model owns a stronger capability to distinguish salient objects.

In conclusion, our proposed method outperforms others on the whole. As the only two models which solve the SOR problem directly, our network exceeds ASRNet among all metrics and brings up significant improvements.

\paragraph{Qualitative Evaluation} We show visualization results in Fig. \ref{vis} to make the qualitative comparison. In the 1st column, due to end-to-end training and optimizing detection branch and SOR branch jointly, our network achieves better detection result in rank 1 class. Note columns from 2 to 4, objects in the center are more salient and deserve higher ranking orders in these images. ASRNet does not take positional information into account and gets wrong ranking orders, while our method gets correct ranking orders. In the 5th column, we can see the bench is an attractive object. However, affected by human existence, the saliency of the bench should be lower than human. ASRNet does not consider feature interaction and thus gets the wrong ranking order. These visualization results show that our model owns much more capability to capture relative saliency information with position embedding and feature interaction.

\subsection{Ablation Study}
\paragraph{End-to-End Training Strategy} ASRNet~\cite{siris2020inferring} adopts two-stage training, while our method employs end-to-end training. To make comprehensive comparisons, we also take a two-stage training strategy on our network. In the first stage, we only train the backbone and detection branch and freeze weights of the SOR branch. The target in this stage is to train a reliable detector. In the second stage, we freeze all weights of the backbone and detection branch and only train the SOR branch. In this stage, we only pay attention to ranking order. The results are shown in Table \ref{end-to-end}. Because of joint optimization, our end-to-end model is better than our two-stage model. It is noteworthy that even using a two-stage training strategy, our network still achieves better performance than ASRNet, which indicates position embedding and feature interaction stage do work and contribute to ranking salient objects.

\begin{table}[!tbh]
\begin{center}

\begin{tabular}{lccc}
\hline
Method & MAE$\downarrow$ & SOR$\uparrow$ & \#Imgs used$\uparrow$ \\
\hline
ASRNet & 0.101 & 0.792 & 2365 \\
Ours (two-stage) & 0.082 & 0.835 & 2369 \\      
Ours (end-to-end)  & \textbf{0.081} & \textbf{0.841} & \textbf{2371}       \\
\hline
\end{tabular}
\end{center}

\caption{The comparison between ASRNet and our model by using two-stage training and end-to-end training.}
\label{end-to-end}
\end{table}

\paragraph{Cooperation with other detectors} Our proposed Multi-Task learning framework can cooperate with multiple detection methods obtaining comparable performance. In other words, the PPA module can be considered as a plug-in module that compatible with region-based prediction tasks such as instance segmentation. To illustrate the effectiveness, we conduct experiments on another prevalent instance segmentation method Mask-RCNN~\cite{he2017mask}, and present the results in Table \ref{Plug-in}. In MaskRCNN experiments, we use ResNet-50/101~\cite{he2016deep} as backbone separately, and they are denoted as MaskRCNN-50/101. In CenterMask experiments, we use VoV-39/57~\cite{lee2019energy} as the backbone, and they are denoted as CenterMask-39/57. As we can see, even though we use different detection methods, all these models still achieve better performance than previous work. It indicates our PPA module can serve as a plug-in module and make a significant improvement in the SOR task.

\begin{table}[!htb]
\begin{center}
\begin{tabular}{lccc}
\hline
Backbone & MAE$\downarrow$ & SOR$\uparrow$ & \#Imgs used$\uparrow$ \\
\hline
MaskRCNN-50~\cite{he2017mask} & 0.097 & 0.817 & 2354 \\
MaskRCNN-101~\cite{he2017mask} & 0.094 & 0.826 & 2366 \\
CenterMask-39~\cite{lee2020centermask}  & \textbf{0.081} & 0.841 & 2371  \\
CenterMask-57~\cite{lee2020centermask}  & 0.085 & \textbf{0.848} & \textbf{2376}  \\
\hline
\end{tabular}
\end{center}
\caption{Experiments on different instance segmentation methods with the proposed PPA module as plugin.}
\label{Plug-in}
\end{table}

\paragraph{Analysis on each stage in PPA} To illustrate both of position embedding stage and feature interaction stage can help to make improvements in the SOR task, we investigate the effectiveness of them respectively. The results are shown in Table \ref{component analysis}.
We design a simple SOR branch as the baseline. Proposals' features first pass four convolution layers. Then these features are flattened to 1-D vectors to obtain \textit{visual tokens} (without position embedding). Finally these \textit{visual tokens} are sent to a fully connected layer to predict ranking orders. To make a comparison, we utilize the position embedding stage but without using the feature interaction stage. The result of this setting is shown in the 2nd row in Table \ref{component analysis}. The experiment in the 3rd row takes the opposite setting. The experiment in the 4th row uses both the position embedding stage and the feature interaction stage. From the results, we can see two stages in PPA can improve the performance in three metrics. These results suggest that both position embedding and feature interaction play essential roles in the SOR task.

\begin{table}

\begin{center}
\begin{tabular}{lccc}
\hline
Method & MAE$\downarrow$ & SOR$\uparrow$ & \#Imgs used$\uparrow$ \\
\hline
Baseline & 0.104 & 0.830 & 2176 \\
Baseline+pos & 0.095 & 0.836 & 2344 \\
Baseline+attention & 0.088 & 0.839 & 2365 \\      
Baseline+attention+pos   & \textbf{0.081} & \textbf{0.841} & \textbf{2371} \\       
\hline
\end{tabular}
\end{center}
\caption{Experiments of the proposed components with the same backbone and detection branch.}
\label{component analysis}

\end{table}

\paragraph{Position Embedding} As mentioned in Section \ref{sec:pos_emb}, we try different position embedding schemes to make comprehensive comparisons. The results are shown in Table \ref{Position Embedding}. The first two methods only utilize bounding box information which follows the scheme described in Fig. \ref{pos} (a). The main difference between the two methods is the first only uses center coordinates information while the second uses both center coordinates and scales information. The third and the fourth methods
follow the scheme described in Fig. \ref{pos} (b). The main difference between the two methods is the previous method only uses center positions as learnable embeddings, and the other method uses both center positions and scales as learnable embeddings. The quantitative number $q$ of position is set to 8, and the quantitative number $q$ of scale is set to 4. The last method follows the scheme described in Fig. \ref{pos} (c) and achieves the best performance. It denotes the importance of making full use of absolute positional information directly and fusing other objects' features. It is also noteworthy that according to the scheme (a) and scheme (b) results, we can find that scale information helps to achieve better performance, which also confirms our previous points.

\begin{table}[!tbh]
\begin{center}
\begin{tabular}{lccc}
\hline
Method & MAE$\downarrow$ & SOR$\uparrow$ & \#Imgs used$\uparrow$ \\
\hline
$C_x, C_y$ & 0.092 & 0.835 & 2366 \\ 
$C_x, C_y, w, h$ & 0.088 & 0.836 & 2370 \\
\hline
Learnable pos & 0.082 & 0.821 & 2368 \\
Learnable pos and scale & 0.083 & 0.834 & 2370 \\
\hline
Ours (PPA)   & \textbf{0.081} & \textbf{0.841} & \textbf{2371}        \\
\hline
\end{tabular}
\end{center}
\caption{The comparison between different position embedding schemes by using the same backbone and detection branch.}
\label{Position Embedding}

\end{table}

\section{Conclusion}
In this paper, we propose the first end-to-end framework of the Salient Object Ranking task and solve it in a multi-task learning fashion. The framework performs instance segmentation and salient object ranking simultaneously. We also propose a Position-Preserved Attention (PPA) module tailored for the SOR branch. This module utilizes positional information and mutual information effectively, which are essential factors of the SOR task. Our method outperforms the state-of-the-art method significantly on the ASR dataset.
\newpage
{\small
\bibliographystyle{ieee_fullname}
\bibliography{egbib}

\begin{thebibliography}{10}\itemsep=-1pt

\bibitem{achanta2008salient}
Radhakrishna Achanta, Francisco Estrada, Patricia Wils, and Sabine
  S{\"u}sstrunk.
\newblock Salient region detection and segmentation.
\newblock In {\em International conference on computer vision systems}, pages
  66--75. Springer, 2008.

\bibitem{ba2016layer}
Jimmy~Lei Ba, Jamie~Ryan Kiros, and Geoffrey~E Hinton.
\newblock Layer normalization.
\newblock {\em arXiv preprint arXiv:1607.06450}, 2016.

\bibitem{borji2019saliency}
Ali Borji.
\newblock Saliency prediction in the deep learning era: Successes and
  limitations.
\newblock {\em IEEE transactions on pattern analysis and machine intelligence},
  2019.

\bibitem{bottou2010large}
L{\'e}on Bottou.
\newblock Large-scale machine learning with stochastic gradient descent.
\newblock In {\em Proceedings of COMPSTAT'2010}, pages 177--186. Springer,
  2010.

\bibitem{brown2020language}
Tom~B Brown, Benjamin Mann, Nick Ryder, Melanie Subbiah, Jared Kaplan, Prafulla
  Dhariwal, Arvind Neelakantan, Pranav Shyam, Girish Sastry, Amanda Askell,
  et~al.
\newblock Language models are few-shot learners.
\newblock {\em arXiv preprint arXiv:2005.14165}, 2020.

\bibitem{carion2020end}
Nicolas Carion, Francisco Massa, Gabriel Synnaeve, Nicolas Usunier, Alexander
  Kirillov, and Sergey Zagoruyko.
\newblock End-to-end object detection with transformers.
\newblock In {\em European Conference on Computer Vision}, pages 213--229.
  Springer, 2020.

\bibitem{chen2016automatic}
Jiansheng Chen, Gaocheng Bai, Shaoheng Liang, and Zhengqin Li.
\newblock Automatic image cropping: A computational complexity study.
\newblock In {\em Proceedings of the IEEE Conference on Computer Vision and
  Pattern Recognition}, pages 507--515, 2016.

\bibitem{chen2020generative}
Mark Chen, Alec Radford, Rewon Child, Jeffrey Wu, Heewoo Jun, David Luan, and
  Ilya Sutskever.
\newblock Generative pretraining from pixels.
\newblock In {\em International Conference on Machine Learning}, pages
  1691--1703. PMLR, 2020.

\bibitem{cheng2014global}
Ming-Ming Cheng, Niloy~J Mitra, Xiaolei Huang, Philip~HS Torr, and Shi-Min Hu.
\newblock Global contrast based salient region detection.
\newblock {\em IEEE transactions on pattern analysis and machine intelligence},
  37(3):569--582, 2014.

\bibitem{devlin2018bert}
Jacob Devlin, Ming-Wei Chang, Kenton Lee, and Kristina Toutanova.
\newblock Bert: Pre-training of deep bidirectional transformers for language
  understanding.
\newblock {\em arXiv preprint arXiv:1810.04805}, 2018.

\bibitem{dosovitskiy2020image}
Alexey Dosovitskiy, Lucas Beyer, Alexander Kolesnikov, Dirk Weissenborn,
  Xiaohua Zhai, Thomas Unterthiner, Mostafa Dehghani, Matthias Minderer, Georg
  Heigold, Sylvain Gelly, et~al.
\newblock An image is worth 16x16 words: Transformers for image recognition at
  scale.
\newblock {\em arXiv preprint arXiv:2010.11929}, 2020.

\bibitem{fan2019s4net}
Ruochen Fan, Ming-Ming Cheng, Qibin Hou, Tai-Jiang Mu, Jingdong Wang, and
  Shi-Min Hu.
\newblock S4net: Single stage salient-instance segmentation.
\newblock In {\em Proceedings of the IEEE/CVF Conference on Computer Vision and
  Pattern Recognition}, pages 6103--6112, 2019.

\bibitem{goyal2017accurate}
Priya Goyal, Piotr Doll{\'a}r, Ross Girshick, Pieter Noordhuis, Lukasz
  Wesolowski, Aapo Kyrola, Andrew Tulloch, Yangqing Jia, and Kaiming He.
\newblock Accurate, large minibatch sgd: Training imagenet in 1 hour.
\newblock {\em arXiv preprint arXiv:1706.02677}, 2017.

\bibitem{he2017mask}
Kaiming He, Georgia Gkioxari, Piotr Doll{\'a}r, and Ross Girshick.
\newblock Mask r-cnn.
\newblock In {\em Proceedings of the IEEE international conference on computer
  vision}, pages 2961--2969, 2017.

\bibitem{he2016deep}
Kaiming He, Xiangyu Zhang, Shaoqing Ren, and Jian Sun.
\newblock Deep residual learning for image recognition.
\newblock In {\em Proceedings of the IEEE conference on computer vision and
  pattern recognition}, pages 770--778, 2016.

\bibitem{he2017delving}
Shengfeng He, Jianbo Jiao, Xiaodan Zhang, Guoqiang Han, and Rynson~WH Lau.
\newblock Delving into salient object subitizing and detection.
\newblock In {\em Proceedings of the IEEE International Conference on Computer
  Vision}, pages 1059--1067, 2017.

\bibitem{he2021transreid}
Shuting He, Hao Luo, Pichao Wang, Fan Wang, Hao Li, and Wei Jiang.
\newblock Transreid: Transformer-based object re-identification.
\newblock {\em arXiv preprint arXiv:2102.04378}, 2021.

\bibitem{hendrycks2016gaussian}
Dan Hendrycks and Kevin Gimpel.
\newblock Gaussian error linear units (gelus).
\newblock {\em arXiv preprint arXiv:1606.08415}, 2016.

\bibitem{hou2017deeply}
Qibin Hou, Ming-Ming Cheng, Xiaowei Hu, Ali Borji, Zhuowen Tu, and Philip~HS
  Torr.
\newblock Deeply supervised salient object detection with short connections.
\newblock In {\em Proceedings of the IEEE conference on computer vision and
  pattern recognition}, pages 3203--3212, 2017.

\bibitem{hu2017deep}
Ping Hu, Bing Shuai, Jun Liu, and Gang Wang.
\newblock Deep level sets for salient object detection.
\newblock In {\em Proceedings of the IEEE conference on computer vision and
  pattern recognition}, pages 2300--2309, 2017.

\bibitem{huang2020hand}
Lin Huang, Jianchao Tan, Ji Liu, and Junsong Yuan.
\newblock Hand-transformer: Non-autoregressive structured modeling for 3d hand
  pose estimation.
\newblock In {\em European Conference on Computer Vision}, pages 17--33.
  Springer, 2020.

\bibitem{islam2018revisiting}
Md~Amirul Islam, Mahmoud Kalash, and Neil~DB Bruce.
\newblock Revisiting salient object detection: Simultaneous detection, ranking,
  and subitizing of multiple salient objects.
\newblock In {\em Proceedings of the IEEE Conference on Computer Vision and
  Pattern Recognition}, pages 7142--7150, 2018.

\bibitem{itti1998model}
Laurent Itti, Christof Koch, and Ernst Niebur.
\newblock A model of saliency-based visual attention for rapid scene analysis.
\newblock {\em IEEE Transactions on pattern analysis and machine intelligence},
  20(11):1254--1259, 1998.

\bibitem{jiang2015salicon}
Ming Jiang, Shengsheng Huang, Juanyong Duan, and Qi Zhao.
\newblock Salicon: Saliency in context.
\newblock In {\em Proceedings of the IEEE conference on computer vision and
  pattern recognition}, pages 1072--1080, 2015.

\bibitem{7780764}
B. {Lai} and X. {Gong}.
\newblock Saliency guided dictionary learning for weakly-supervised image
  parsing.
\newblock In {\em 2016 IEEE Conference on Computer Vision and Pattern
  Recognition (CVPR)}, pages 3630--3639, 2016.

\bibitem{lee2016deep}
Gayoung Lee, Yu-Wing Tai, and Junmo Kim.
\newblock Deep saliency with encoded low level distance map and high level
  features.
\newblock In {\em Proceedings of the IEEE Conference on Computer Vision and
  Pattern Recognition}, pages 660--668, 2016.

\bibitem{lee2019energy}
Youngwan Lee, Joong-won Hwang, Sangrok Lee, Yuseok Bae, and Jongyoul Park.
\newblock An energy and gpu-computation efficient backbone network for
  real-time object detection.
\newblock In {\em Proceedings of the IEEE/CVF Conference on Computer Vision and
  Pattern Recognition Workshops}, pages 0--0, 2019.

\bibitem{lee2020centermask}
Youngwan Lee and Jongyoul Park.
\newblock Centermask: Real-time anchor-free instance segmentation.
\newblock In {\em Proceedings of the IEEE/CVF conference on computer vision and
  pattern recognition}, pages 13906--13915, 2020.

\bibitem{li2016deep}
Guanbin Li and Yizhou Yu.
\newblock Deep contrast learning for salient object detection.
\newblock In {\em Proceedings of the IEEE conference on computer vision and
  pattern recognition}, pages 478--487, 2016.

\bibitem{li2014secrets}
Yin Li, Xiaodi Hou, Christof Koch, James~M Rehg, and Alan~L Yuille.
\newblock The secrets of salient object segmentation.
\newblock In {\em Proceedings of the IEEE conference on computer vision and
  pattern recognition}, pages 280--287, 2014.

\bibitem{lin2014microsoft}
Tsung-Yi Lin, Michael Maire, Serge Belongie, James Hays, Pietro Perona, Deva
  Ramanan, Piotr Doll{\'a}r, and C~Lawrence Zitnick.
\newblock Microsoft coco: Common objects in context.
\newblock In {\em European conference on computer vision}, pages 740--755.
  Springer, 2014.

\bibitem{liu2016dhsnet}
Nian Liu and Junwei Han.
\newblock Dhsnet: Deep hierarchical saliency network for salient object
  detection.
\newblock In {\em Proceedings of the IEEE conference on computer vision and
  pattern recognition}, pages 678--686, 2016.

\bibitem{liu2010learning}
Tie Liu, Zejian Yuan, Jian Sun, Jingdong Wang, Nanning Zheng, Xiaoou Tang, and
  Heung-Yeung Shum.
\newblock Learning to detect a salient object.
\newblock {\em IEEE Transactions on Pattern analysis and machine intelligence},
  33(2):353--367, 2010.

\bibitem{luo2017non}
Zhiming Luo, Akshaya Mishra, Andrew Achkar, Justin Eichel, Shaozi Li, and
  Pierre-Marc Jodoin.
\newblock Non-local deep features for salient object detection.
\newblock In {\em Proceedings of the IEEE Conference on computer vision and
  pattern recognition}, pages 6609--6617, 2017.

\bibitem{ma2003contrast}
Yu-Fei Ma and Hong-Jiang Zhang.
\newblock Contrast-based image attention analysis by using fuzzy growing.
\newblock In {\em Proceedings of the eleventh ACM international conference on
  Multimedia}, pages 374--381, 2003.

\bibitem{najibi2018towards}
Mahyar Najibi, Fan Yang, Qiaosong Wang, and Robinson Piramuthu.
\newblock Towards the success rate of one: Real-time unconstrained salient
  object detection.
\newblock In {\em 2018 IEEE Winter Conference on Applications of Computer
  Vision (WACV)}, pages 1432--1441. IEEE, 2018.

\bibitem{qin2019basnet}
Xuebin Qin, Zichen Zhang, Chenyang Huang, Chao Gao, Masood Dehghan, and Martin
  Jagersand.
\newblock Basnet: Boundary-aware salient object detection.
\newblock In {\em Proceedings of the IEEE/CVF Conference on Computer Vision and
  Pattern Recognition}, pages 7479--7489, 2019.

\bibitem{ren2015faster}
Shaoqing Ren, Kaiming He, Ross Girshick, and Jian Sun.
\newblock Faster r-cnn: Towards real-time object detection with region proposal
  networks.
\newblock {\em arXiv preprint arXiv:1506.01497}, 2015.

\bibitem{siris2020inferring}
Avishek Siris, Jianbo Jiao, Gary~KL Tam, Xianghua Xie, and Rynson~WH Lau.
\newblock Inferring attention shift ranks of objects for image saliency.
\newblock In {\em Proceedings of the IEEE/CVF Conference on Computer Vision and
  Pattern Recognition}, pages 12133--12143, 2020.

\bibitem{tighe2013finding}
Joseph Tighe and Svetlana Lazebnik.
\newblock Finding things: Image parsing with regions and per-exemplar
  detectors.
\newblock In {\em Proceedings of the IEEE conference on computer vision and
  pattern recognition}, pages 3001--3008, 2013.

\bibitem{vaswani2017attention}
Ashish Vaswani, Noam Shazeer, Niki Parmar, Jakob Uszkoreit, Llion Jones,
  Aidan~N Gomez, Lukasz Kaiser, and Illia Polosukhin.
\newblock Attention is all you need.
\newblock {\em arXiv preprint arXiv:1706.03762}, 2017.

\bibitem{wang2020max}
Huiyu Wang, Yukun Zhu, Hartwig Adam, Alan Yuille, and Liang-Chieh Chen.
\newblock Max-deeplab: End-to-end panoptic segmentation with mask transformers.
\newblock {\em arXiv preprint arXiv:2012.00759}, 2020.

\bibitem{wang2016saliency}
Linzhao Wang, Lijun Wang, Huchuan Lu, Pingping Zhang, and Xiang Ruan.
\newblock Saliency detection with recurrent fully convolutional networks.
\newblock In {\em European conference on computer vision}, pages 825--841.
  Springer, 2016.

\bibitem{wang2019learning}
Qiang Wang, Bei Li, Tong Xiao, Jingbo Zhu, Changliang Li, Derek~F Wong, and
  Lidia~S Chao.
\newblock Learning deep transformer models for machine translation.
\newblock {\em arXiv preprint arXiv:1906.01787}, 2019.

\bibitem{wang2017stagewise}
Tiantian Wang, Ali Borji, Lihe Zhang, Pingping Zhang, and Huchuan Lu.
\newblock A stagewise refinement model for detecting salient objects in images.
\newblock In {\em Proceedings of the IEEE International Conference on Computer
  Vision}, pages 4019--4028, 2017.

\bibitem{wang2021salient}
Wenguan Wang, Qiuxia Lai, Huazhu Fu, Jianbing Shen, Haibin Ling, and Ruigang
  Yang.
\newblock Salient object detection in the deep learning era: An in-depth
  survey.
\newblock {\em IEEE Transactions on Pattern Analysis and Machine Intelligence},
  2021.

\bibitem{wang2019ranking}
Zheng Wang, Xinyu Yan, Yahong Han, and Meijun Sun.
\newblock Ranking video salient object detection.
\newblock In {\em Proceedings of the 27th ACM International Conference on
  Multimedia}, pages 873--881, 2019.

\bibitem{wu2019cascaded}
Zhe Wu, Li Su, and Qingming Huang.
\newblock Cascaded partial decoder for fast and accurate salient object
  detection.
\newblock In {\em Proceedings of the IEEE/CVF Conference on Computer Vision and
  Pattern Recognition}, pages 3907--3916, 2019.

\bibitem{wu2019stacked}
Zhe Wu, Li Su, and Qingming Huang.
\newblock Stacked cross refinement network for edge-aware salient object
  detection.
\newblock In {\em Proceedings of the IEEE/CVF International Conference on
  Computer Vision}, pages 7264--7273, 2019.

\bibitem{xu2015show}
Kelvin Xu, Jimmy Ba, Ryan Kiros, Kyunghyun Cho, Aaron Courville, Ruslan
  Salakhudinov, Rich Zemel, and Yoshua Bengio.
\newblock Show, attend and tell: Neural image caption generation with visual
  attention.
\newblock In {\em International conference on machine learning}, pages
  2048--2057. PMLR, 2015.

\bibitem{yao2018exploring}
Ting Yao, Yingwei Pan, Yehao Li, and Tao Mei.
\newblock Exploring visual relationship for image captioning.
\newblock In {\em Proceedings of the European conference on computer vision
  (ECCV)}, pages 684--699, 2018.

\bibitem{zeng2020learning}
Yanhong Zeng, Jianlong Fu, and Hongyang Chao.
\newblock Learning joint spatial-temporal transformations for video inpainting.
\newblock In {\em European Conference on Computer Vision}, pages 528--543.
  Springer, 2020.

\bibitem{zhang2016unconstrained}
Jianming Zhang, Stan Sclaroff, Zhe Lin, Xiaohui Shen, Brian Price, and Radomir
  Mech.
\newblock Unconstrained salient object detection via proposal subset
  optimization.
\newblock In {\em Proceedings of the IEEE conference on computer vision and
  pattern recognition}, pages 5733--5742, 2016.

\bibitem{zhang2017amulet}
Pingping Zhang, Dong Wang, Huchuan Lu, Hongyu Wang, and Xiang Ruan.
\newblock Amulet: Aggregating multi-level convolutional features for salient
  object detection.
\newblock In {\em Proceedings of the IEEE International Conference on Computer
  Vision}, pages 202--211, 2017.

\bibitem{zhang2017learning}
Pingping Zhang, Dong Wang, Huchuan Lu, Hongyu Wang, and Baocai Yin.
\newblock Learning uncertain convolutional features for accurate saliency
  detection.
\newblock In {\em Proceedings of the IEEE International Conference on computer
  vision}, pages 212--221, 2017.

\bibitem{zhu2021horizontal}
Tun Zhu, Daoxin Zhang, Tianran Wang, Xiaolong Jiang, Jiawei Li, Yao Hu, and
  Jianke Zhu.
\newblock Horizontal-to-vertical video conversion.
\newblock {\em arXiv preprint arXiv:2101.04051}, 2021.

\end{thebibliography}
}

\end{document}